\documentclass[acmsmall,nonacm]{acmart}

\usepackage{color}
\usepackage{booktabs}
\usepackage{textcomp}
\usepackage{graphicx, subcaption}
\usepackage{diagbox}
\usepackage{multirow}
\usepackage{adjustbox}
\usepackage[labelformat=simple]{subcaption}

\DeclareCaptionLabelFormat{subcaptionlabel}{\normalfont(\textbf{#2}\normalfont)}
\usepackage{tabularx}
\usepackage{calc}
\usepackage{arydshln}
\usepackage{makecell}

\newcolumntype{Y}{>{\centering\arraybackslash}X}
\usepackage{array}
\usepackage{enumitem}
\usepackage{bm}
\newcolumntype{?}{!{\vrule width 1pt}}

\AtBeginDocument{%
  \providecommand\BibTeX{{%
    \normalfont B\kern-0.5em{\scshape i\kern-0.25em b}\kern-0.8em\TeX}}}

\begin{document}

\title{Assessing Distribution Shift in Human Activity Recognition for Domain Generalization}

\author{Rebecca Adaimi}
\email{rebecca.adaimi@utexas.edu}
\orcid{}
\affiliation{%
  \institution{The University of Texas at Austin}
  \streetaddress{}
  \city{Austin}
  \state{Texas}
  \country{USA}
  \postcode{}
}

\author{Edison Thomaz}
\email{ethomaz@utexas.edu}
\orcid{}
\affiliation{%
  \institution{The University of Texas at Austin}
  \streetaddress{}
  \city{Austin}
  \state{Texas}
  \country{USA}
  \postcode{}
}

\renewcommand{\shortauthors}{Adaimi, et al.}

\begin{abstract}
    While the field of Human Activity Recognition (HAR) continues to draw interest from researchers and advance in important ways, some key challenges remain. One of the most difficult aspects of building HAR models that show good performance in real-world settings is dealing with data diversity from device and sensor heterogeneity, and contextual changes that are intrinsic to real-world applications. While data diversity in HAR has been well-acknowledged in the literature,  there remains a gap in understanding the effect of various types of distribution shifts on HAR models and the domain generalization problem that arises. Towards that end, this paper systematically evaluates 4 different types of distribution shifts, including variations in device type, sensor placement, sampling rate, and user behavior. Quantifying their effects, we illustrate that diversity shifts predominantly define all types of shifts, indicating the existence of unique features that are not shared across different domains. We then introduce a uniform HAR-based distribution shift benchmarks and conduct a comprehensive evaluation of up to 28 domain generalization methods. Our analysis exposes the limitations of current domain generalization algorithms in achieving model generalizability, marginally outperforming the empirical risk minimization baseline. This work represents the first systematic exploration of domain generalization and adaptation concerning specific distribution shifts in sensor-based HAR, offering an open-source benchmark platform and datasets to spur further research.
     
     % For each, we quantify their impact on the data and evaluate 28 domain generalization methods. Our experiments reveal the limitations of current domain generalization algorithms in achieving model generalizability in sensor-based human activity. This work marks the first attempt to systematically explore domain generalization and adaptation concerning specific distribution shifts in sensor-based HAR, providing an open-sourced benchmark platform and datasets to drive new research.  

\end{abstract}


\received{20 February 2007}
\received[revised]{12 March 2009}
\received[accepted]{5 June 2009}

\maketitle

\section{Introduction}

Over the past decade, thanks to advances in smartphones, sensors and powerful machine learning methods, interest in the field of Human Activity Recognition (HAR) has gained significant strength. This growing area of research has been driven by a plethora of opportunities in mobile health \cite{10.1145/3301275.3302315,san2020eating}, disorder diagnosis \cite{cakmak2020using}, personal assistance, smart environments \cite{OkGoogle}, and numerous other domains. HAR, once primarily limited to controlled laboratory environments, has now transcended these boundaries to operate in the complex and dynamic contexts of our everyday routines. However, despite much progress, a vast majority of HAR methods is still largely based on the assumption that training and test data used for modeling have similar distributions. This assumption is often not valid in real-world applications due to ubiquitous distribution shifts introduced by variations in device types, sensor placements, and contextual environments. As a result, state-of-the-art methods continue to face challenges in bridging the gap between controlled validation conditions and the intricate complexities of the real world \cite{10.1145/3534582}.

From a machine learning perspective, the challenge of data diversity in HAR models can be formulated as a \textit{domain shift} problem. To tackle the aforementioned challenges, it is crucial to assess the generalizability of a model on novel data. An effective approach frequently employed to address domain shift issues is transfer learning or domain adaptation \cite{gjoreski2019cross,10.1145/2971763.2971764,wang2018deep,zhao2020local}, which involves utilizing knowledge acquired from other available domains during the training phase, and then fine-tuning the model on a new target domain to minimize the distribution gap and improve overall performance. Nevertheless, a primary assumption for such techniques is the availability or accessibility of the target data, which is often unfeasible in practical applications. This is particularly true in the realm of sensor-based activity data, where it is not practical to amass sensor data for all potential distribution changes. Thus, domain generalization is a research topic that focuses on the problem of building a model that generalizes well to unseen data without access to the new data during training. 

As noted earlier, the challenge of generalization in human activity recognition (HAR) can arise from various factors, such as changes in device type, sensor heterogeneity in terms of sampling rate or sensor placement, environments, and activity patterns, among others. While these challenges have been recognized in previous research, it is still unclear to the scientific community which of these factors specifically makes domain generalization in HAR particularly difficult. This paper takes the initial step in conducting a methodical evaluation of the generalizability of HAR models across these distribution changes. 
% We urge researchers to include this critical step in their model analysis pipeline to further advance the understanding of HAR model performance in real-world scenarios.

% To perform this analysis, we first investigate the different causes of distribution shifts, focusing on sensor-related changes (e.g. device type, sensor placement and sampling rate), user-related changes (e.g. skill improvement), and environment-related changed (e.g. controlled vs. in-the-wild). Taking inspiration from the work of Ye \textit{et al.} \cite{ye2022ood}, we quantify each distribution shift using a two-dimensional spectrum, which includes measurements of diversity shift and correlation shift. Second, there have been significant advancements in domain generalization methods within the machine learning community, particularly in the fields of computer vision and natural language processing \cite{zhou2022domain}. However, the problem of domain generalization in HAR has received limited attention.

In our investigation of the problem space, we structured our analysis to address two primary research questions:
\begin{enumerate}[label=\arabic*.]
    \item \textbf{\textit{Quantifying Distribution Shifts.}} Focusing on sensor-based HAR data, what are the different causes of distribution shifts, such as sensor-related changes (e.g. device type, sensor placement and sampling rate) and user-related changes (e.g. skill improvement)? And can we represent and quantify each shift to better understand its influence on the data? Taking inspiration from the work of Ye \textit{et al.} \cite{ye2022ood}, we quantify each distribution shift using a two-dimensional spectrum, which includes measurements of diversity shift and correlation shift. Our analysis uncovered that all types of shifts and datasets are primarily characterized by diversity shifts, indicating the existence of unique features that are not shared across different domains. 
    \vspace{10pt}
    \item \textbf{\textit{Domain Generalization.}} From a domain generalization perspective, how do state-of-the-art domain generalization methods perform in the context of these sensor-based distribution shifts? To answer this question, we developed a uniform benchmark platform\footnote{\url{www.to-be-added.com}} and evaluated up to 28 domain generalization methods on 4 types of distribution shifts. Our results show that existing algorithms only marginally improve performance over the empirical risk minimization baseline, emphasizing the challenge of domain generalization across sensor-based sensing heterogeneity and distribution shifts.
    % \item \textcolor{red}{\textbf{\textit{Domain Adaptation.}} From a domain adaptation perspective, can we reduce .... \textcolor{red}{NEED TO FORMULATE THIS STILL}}
\end{enumerate}

To the best of our knowledge, our research is the first to investigate domain generalization and adaptation in the context of specific distribution shifts relevant to sensor-based HAR, with a focus on understanding the challenges each shift poses to model generalizability. It should be noted that our primary objective is not to critique or highlight any particular existing methods, but rather to (1) underscore and demonstrate the significance and deficiency of model generalizability when it comes to sensor-based human activity and (2) present a benchmark platform and datasets for future research.

\section{Related Work}
In this section, we start by providing a brief overview of the recent progress in the area of domain adaptation, with a specific emphasis on HAR. Subsequently, we redirect our focus towards the issue of domain generalization, which serves as the main focus of this paper.

\subsection{Domain Adaptation}

The study of domain adaptation in HAR, also known as cross-domain human activity recognition, is a complex issue that has been explored with different goals. Experts have acknowledged the difficulties involved in transferring knowledge from HAR models that are trained on data from one domain to another \cite{ramasamy2018recent}. Each research endeavor focuses on various sub-problems related to cross-domain HAR, such as cross-sensor-modalities \cite{10.1145/2971763.2971764}, cross-location \cite{wang2018deep}, and cross-environment \cite{7805187}. 

In all of these problems, the primary aim is to modify the model during training so that it can adapt to the target domain and decrease the divergence in distribution. An effective and commonly used approach is the \textit{pretrain-then-finetune} paradigm, also known as transfer learning \cite{10.1145/2971763.2971764}. The idea is using some data from the target domain, the weights of a pretrained model can be finetuned to reduce the distribution shift. This adaptation process has been explored in both supervised \cite{li2021supervised} and unsupervised settings \cite{chang2020systematic}. HDCNN applied feature matching at each layer of the neural network, reducing the Kullback-Leibler divergence and enabling adaptation of the model to a smartwatch using unlabeled data \cite{khan2018scaling}. Chen \textit{et al.} \cite{chen2019motiontransformer} proposed, MotionTransformer, which utilized a confusion maximization approach to translate data from any wearable sensor to match the data collected from a smartphone. Similarly, Akbari \textit{et al.}\cite{akbari2019transferring} utilized variational autoencoders with feature matching to minimize the distribution gap between labeled source and unlabeled target datasets.

As mentioned previously, these techniques operate under the presumption that the target data, either labeled or unlabeled, is present during the training phase. However, this is not always the case in real-world situations. Domain generalization is a learning task that seeks to create a generalizable model capable of generalizing to new domains without prior access, thereby addressing this limitation.

\subsection{Domain Generalization}
The task of building a model that can generalize across multiple domains is a longstanding challenge in the field of machine learning. Recently, several algorithms have been proposed to address this issue, which can be broadly categorized into three groups: (1) data manipulation that utilize data augmentation techniques to support model generalization \cite{zhang2017mixup}, (2) representation learning that aim to develop a generalized feature space \cite{arjovsky2019invariant,ganin2016domain}, and (3) learning strategy that focuses on training procedures \cite{li2018learning,tzeng2015simultaneous,yao2010boosting}. This problem has been extensively studied in other domains such as computer vision \cite{inoue2018cross,li2020domainmedical,li2021simple,wang2020learning,NEURIPS2021_a8f12d94} and natural language processing \cite{li2020domain, nilizadeh2019think}. In order to facilitate research in this field, several benchmark platforms have been developed, such as DomainBed \cite{gulrajani2020search}, DeepDG \cite{wang2022generalizing}, and WILDS \cite{koh2021wilds}. While domain generalization has been extensively studied in the context of image-based tasks, its application to sensor-based data has only recently begun to garner attention. These works include studies on cross-dataset generalization \cite{xu2023globem}, cross-user generalization \cite{qin2022domain}, and cross-position generalization \cite{lu2022semantic}. For instance, Qin \textit{et al.} proposed an adaptive feature fusion approach that learns both domain-invariant and domain-specific features to build a generalizable model across users \cite{qin2022domain}. Another work focused on cross-dataset generalization for depression detection using two multi-year longitudinal behavior datasets \cite{xu2023globem}. They additionally introduce a benchmark platform, GLOBEM, to support future research in this field. Nonetheless, there remains a gap in our understanding of how effectively these algorithms can handle the various types of heterogeneities present in sensor-based data. With its inherent complexity and wide range of possible distribution shifts, HAR can serve as another ideal platform to study the domain generalization abilities of learning algorithms.

\begin{figure}[t]
     \centering

    \begin{subfigure}[b]{0.49\linewidth}
        \includegraphics[width=1.0\columnwidth]{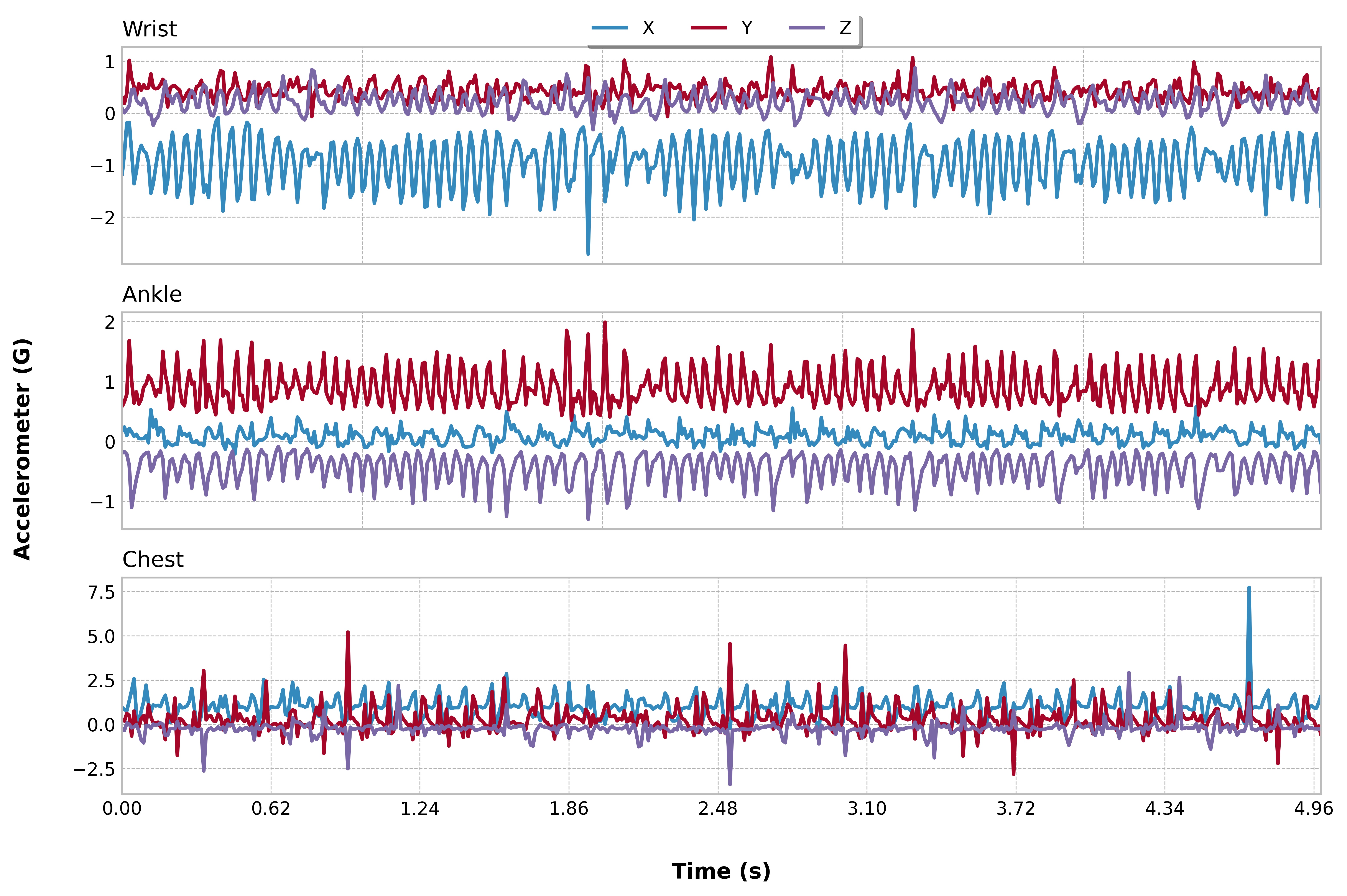}
        \caption{Sensor Location}
        \label{fig:sensor_location}
    \end{subfigure}
    \begin{subfigure}[b]{0.49\linewidth}
        \includegraphics[width=1.0\columnwidth]{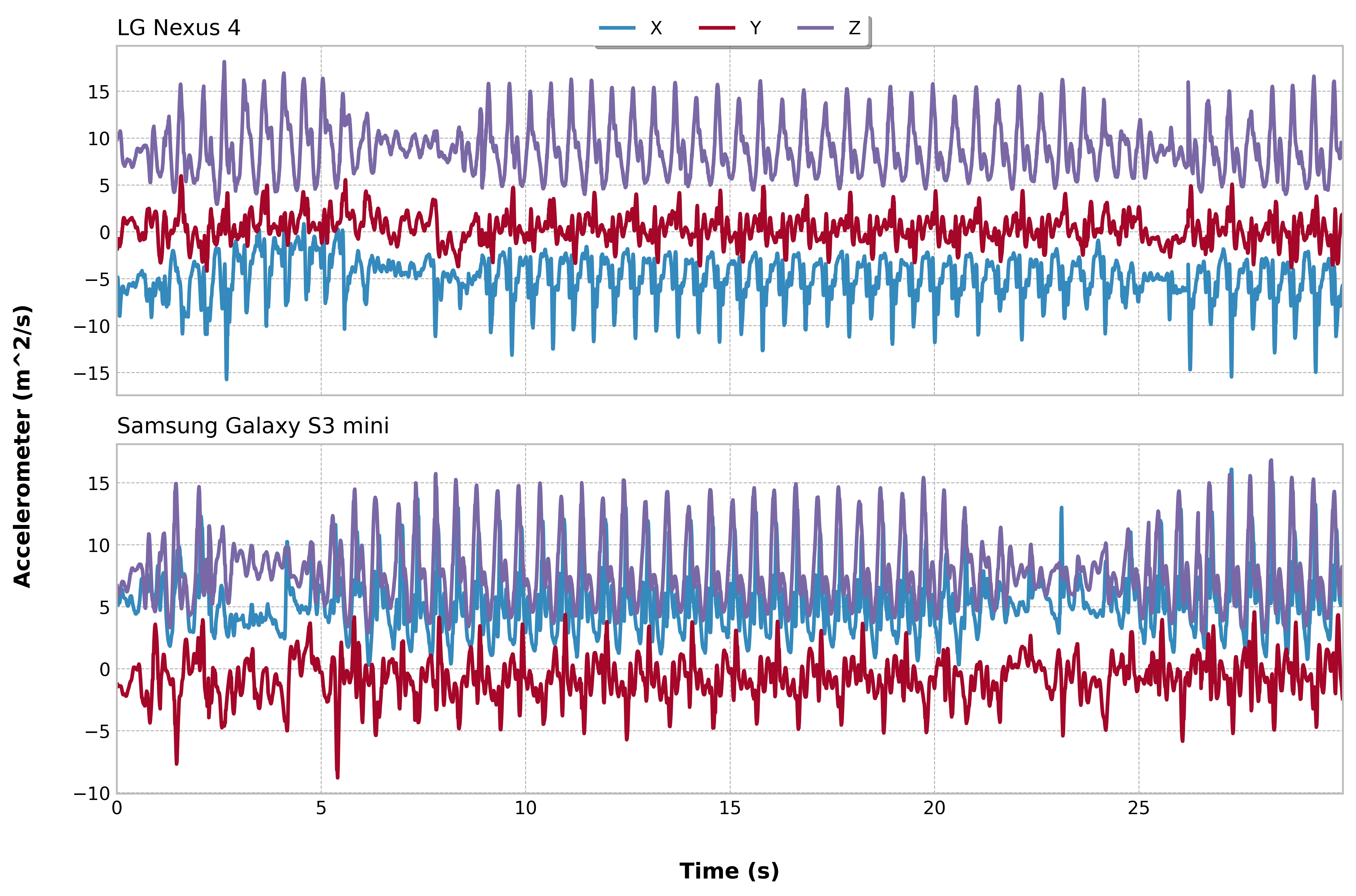}
        \caption{Device Type}
        \label{fig:device_type}
    \end{subfigure}
    \begin{subfigure}[b]{0.49\linewidth}
        \includegraphics[width=1.0\columnwidth]{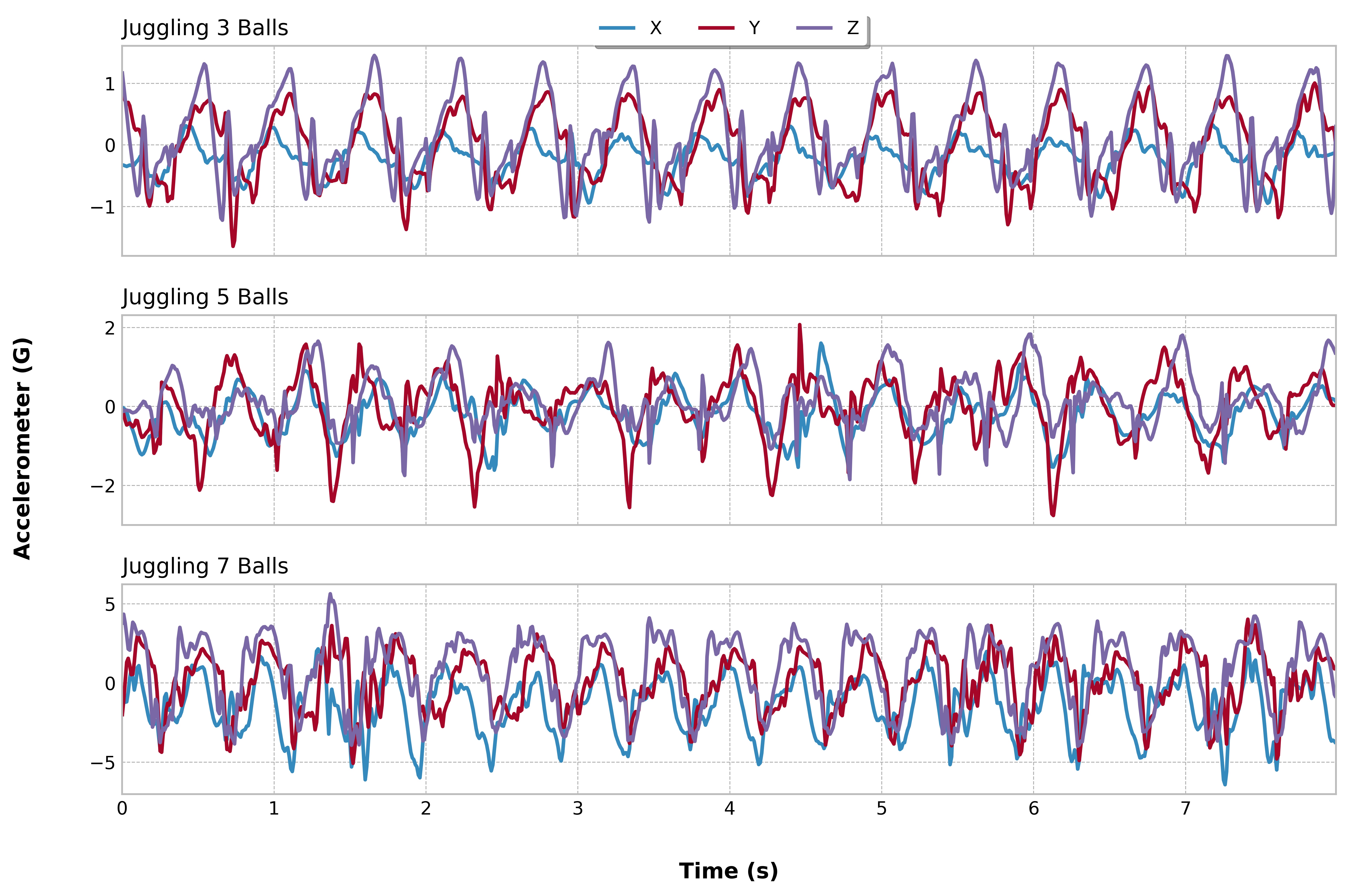}
        \caption{User Behavior}
    \label{fig:user_behavior}
    \end{subfigure}
    \begin{subfigure}[b]{0.49\linewidth}
        \includegraphics[width=1.0\columnwidth]{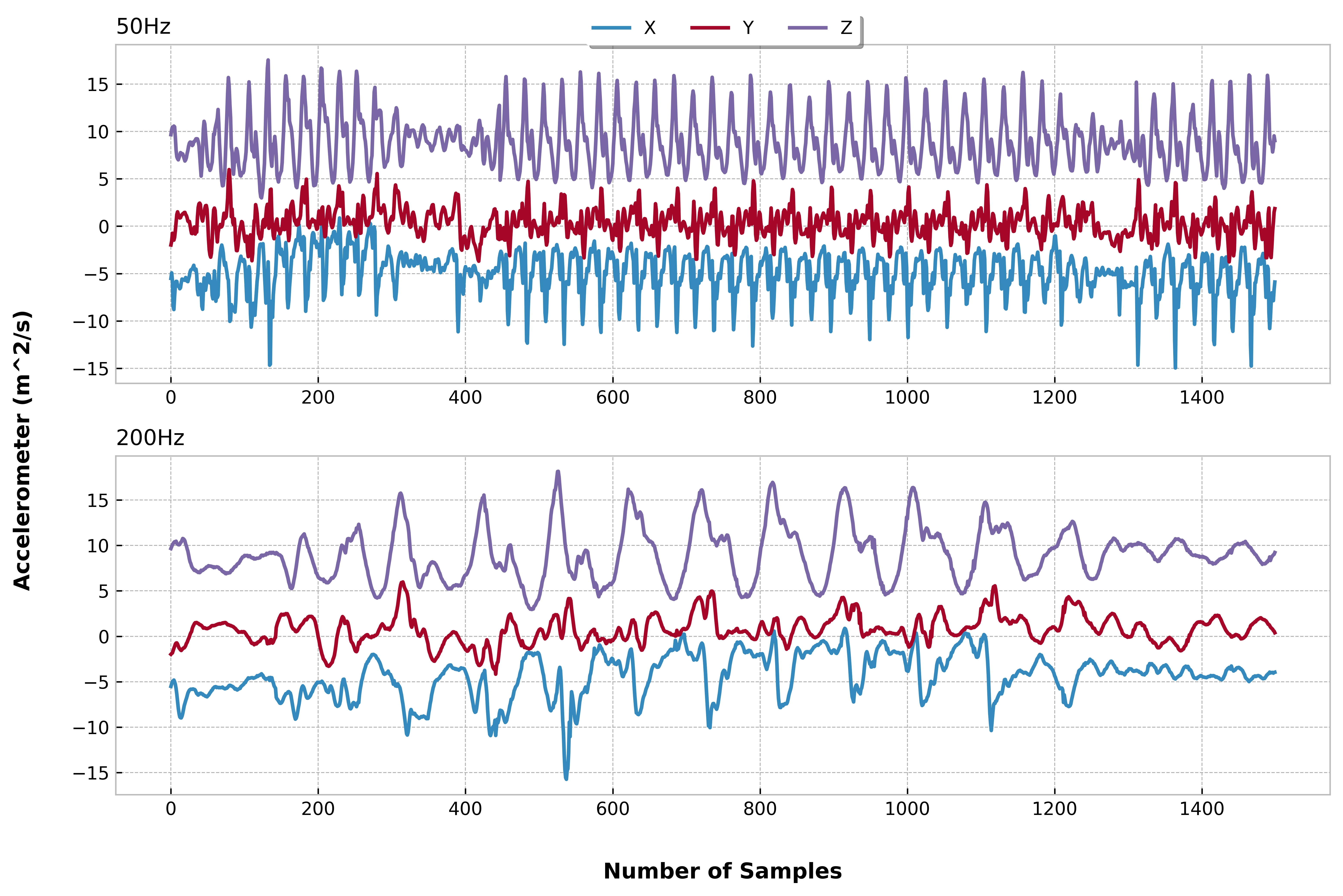}
        \caption{Sampling Rate}
    \label{fig:sampling_rate}
    \end{subfigure}
    \caption{Accelerometer signals captured with varying (a) sensor location (wrist, ankle, chest), (b) Device type (LG Nexus 4 and Samsung Galaxy S3 mini), (c) User Behavior or skill change (juggling 3, 5, and 7 balls), and (d) Sampling Rate (50Hz and 200Hz).}
    \label{fig:shifts}
\end{figure}{}

% \begin{figure}[t]
%     \centering
%     \includegraphics{images/causes_of_shift.png}
%     \caption{plots showing differences in accelerometer for different distribution shifts}
%     \label{fig:shifts}
% \end{figure}

\section{Causes of Distribution Shifts in HAR}
\label{sec:shifts}
Differences in wearable sensing devices can cause changes in the distribution of data, potentially leading to reduced performance of a pre-existing model. These changes can occur due to multiple factors, including but not limited to the type of device used, the location of the sensor, the sampling rate of data collection, and variations in user behavior (Figure \ref{fig:shifts}).

\begin{enumerate}
    \item \textit{\textbf{Device Type}}. Variations in sensor biases across devices can result in changes to the data distribution. For instance, the data collected from accelerometers on different devices such as a Samsung Galaxy S3 mini and LG Nexus 4 can display significant biases and variations, as depicted in Figure \ref{fig:device_type}. We can see slight differences in the accelerometer range between the two devices owing to the differences in the inertial sensors integrated. 
    \item \textit{\textbf{Sensor Location}}. Mobile and wearable devices can be worn or carried by users in various ways due to differences in form factors and personal preferences. For instance, a smartphone can be held in hand or carried in a pocket. As Figure \ref{fig:sensor_location} illustrates, the accelerometer data obtained from wearable devices across three different body locations exhibit significant divergence, indicating a domain shift due to variations in sensor location. We can clearly see differences in the accelerometer range as well as the walking pattern captured from each body location. 
    \item \textit{\textbf{Sampling Rate}}. The maximum sampling rate that a mobile device can support can vary significantly across device models. For instance, while some devices can only attain a maximum sampling rate of 50 Hz, others can reach up to 200 Hz \cite{stisen2015smart}. This disparity can result in a mismatch in sampling rate between the training and test data, which can adversely affect the performance of an activity recognition model. As shown in Figure \ref{fig:sampling_rate}, for a fixed window size, the data captured at 50Hz looks widely different from data captured at 200Hz. 
    \item \textit{\textbf{User Behavior}}. Changes in a user's behavior can also result in distribution shifts. Firstly, the same activity can be performed slightly differently by different users, leading to variations in the data. Moreover, the way a user performs an activity can change over time, particularly in the case of skill-learning tasks. For instance, if a user is learning to juggle, they may start with juggling two balls and then progress to juggling more balls as their skill improves. This progression results in changes in the way the juggling activity is performed over time, as shown in Figure \ref{fig:user_behavior}, which illustrates the change in the accelerometer data when a user juggles 3 balls compared to juggling 5 and 7 balls, in terms of accelerometer range as well as juggling pattern. As expected, the juggling hand movements are faster when juggling 7 balls compared to 3.
\end{enumerate}

% \textit{\textbf{Context Environment}}. 
% In general, human activity recognition models are developed using data collected in a controlled lab environment. However, during deployment in real-world settings, various factors such as participant behavior, device usage, environmental conditions, activity patterns, and context can introduce distribution shifts, making it difficult for the recognition model to perform well. As shown in Figure \ref{fig:context_environment}, accelerometer data captured in a controlled laboratory setting exhibits differences compared to data captured in a real-world setting in terms of noise, user-specific behavior, etc. This highlights the importance of developing models that can effectively handle distribution shifts in order to ensure robust and reliable performance in real-world applications.

\section{Benchmark Experiments}
\label{sec:experiments}

In this section, we introduce a set of HAR-related domain generalization benchmarks representing each type of distribution shift as listed in Section \ref{sec:shifts}. We leverage one or several widely-used publicly available datasets for each of the 3 distribution shifts: (1) \textit{sensor location}, (2)\textit{ sampling rate}, (3) \textit{device type}, and collect a new dataset simulating skill improvement for the \textit{user behavior} shift. As such, we begin by giving a brief description of the datasets used and then explain the steps taken to prepare the datasets for each distribution shift scenario.

\begin{table}[t]
    \centering
    \footnotesize
    \caption{Benchmark Datasets Overview}
    \begin{tabularx}{\textwidth}{@{}XXXcc@{}}%p{2.5cm}|p{3cm}|p{5cm}|c|c@{}}
    \toprule
        \textbf{Distribution Shift} & \textbf{Dataset} & \textbf{Domains} & \textbf{Sampling Rate} & \textbf{Activity}\\
        \midrule
         \multirow{6}{*}{Sensor Location} & \multirow{1}{*}{PAMAP2 \cite{reiss2012creating}} & 3\hspace{.25cm} (wrist, ankle, chest) & \multirow{1}{*}{100 Hz} & \multirow{1}{*}{5}\\
         \cmidrule{2-5}
         & \multirow{2}{*}{DSADS \cite{altun2010human}} & 5 \hspace{.25cm}(torso, right arm, left arm, right leg, left leg) & \multirow{2}{*}{25 Hz} & \multirow{2}{*}{5} \\
         \cmidrule{2-5}
         & \multirow{2}{*}{Opportunity \cite{chavarriaga2013opportunity}} & 6\hspace{.25cm} (right wrist, right knee, left upper arm, left shoe, hip, back) & \multirow{2}{*}{30 Hz} & \multirow{2}{*}{5}\\         
         % & RealWorld \cite{sztyler2016body} & 7\hspace{.25cm} (forearm, thigh, head, upper arm, waist, chest, shin) & 50 Hz & 8 \\
        \midrule
         \multirow{4}{*}{Sampling Rate} & \multirow{1}{*}{PAMAP2} & 3\hspace{.25cm} (10Hz, 50Hz, 100Hz) & \multirow{1}{*}{100 Hz} & \multirow{1}{*}{5}\\
         \cmidrule{2-5}
        % \vspace{.25cm}
         & \multirow{1}{*}{RealWorld \cite{sztyler2016body}} & 3\hspace{.25cm} (10Hz, 25Hz, 50Hz) & \multirow{1}{*}{50 Hz} & \multirow{1}{*}{8}\\
         \cmidrule{2-5}
         & \multirow{1}{*}{HHAR Nexus4 \cite{stisen2015smart}} & 4\hspace{.25cm} (10Hz, 50Hz, 100Hz, 200Hz) & \multirow{1}{*}{200 Hz} & \multirow{1}{*}{6}\\
        \midrule
        % \vspace{.5cm}
         \multirow{2}{*}{Device Type} & \multirow{2}{*}{HHAR \cite{stisen2015smart}} & 4\hspace{.25cm} (Nexus4, Galaxy S3, Galaxy S3 mini, Galaxy S Plus) & \multirow{2}{*}{50-200 Hz} & \multirow{2}{*}{6} \\
        \midrule
        % \vspace{.5cm}
        %  \multirow{1}{*}{Context Environment} & Lab2Wild \cite{bhattacharya2022leveraging} & 3 \hspace{.25cm}(users split into groups) & 50.7 Hz & \\
        % \midrule
        % \vspace{.5cm}
         \multirow{1}{*}{User Behavior} & \multirow{1}{*}{Juggling} & 4 \hspace{.25cm}(juggling 3, 4, 5, and 7 balls) & \multirow{1}{*}{100 Hz} & \multirow{1}{*}{4}\\ 
        \bottomrule
    \end{tabularx}
    \label{tab:datasets}
\end{table}

% \section{Datasets}

When selecting the datasets to accompany our analysis, we focused on datasets (see Table \ref{tab:datasets}) that include at least one of the distribution shifts described in Section \ref{sec:shifts} to create a benchmark for domain generalization. If a public dataset was not readily available, we collected our own dataset to complete the analysis. 

\subsection{Public HAR Datasets}

\subsubsection{PAMAP2} Physical Activity Monitoring dataset \cite{reiss2012creating} contains 18 physical activities captured from 9 subjects sampled at 100 Hz. Each subject wears 3 inertial measurement units (IMU) placed at 3 distinct locations (wrist, ankle, chest) and a heart rate monitor. We use the 3-axis accelerometer data from each IMU. We focus on common full-body activities, and as such cluster activities into: (1) \textit{idle} (lying, sitting, standing), (2) \textit{walking} (walking, ascending/descending stairs, nordic walking), (3) \textit{running}, (4) \textit{cycling}, and (5) \textit{other} (vacuuming, ironing, rope jumping, watching tv, folding laundry, house clearning, playing soccer, car driving, computer work).

\subsubsection{DSADS} Daily and Sports Activities Dataset \cite{altun2010human} collects 19 activities from 8 subjects. Each subject wears 5 on-body sensors that capture triaxial accelerometer, gyroscope, and magnetometer sampled at 25 Hz from 5 locations: torso, right arm, left arm, right leg, and left leg. Similar to PAMAP2, we cluster the activities to form a set of 5 common activities: (1) \textit{idle} (sitting, standing, lying on back/right, standing in elevator), (1) \textit{walking} (ascending/descending stairs, moving in elevator, walking in parking lot/treadmill flat/treadmill at incline, (3) \textit{running}, (4) \textit{cycling} (cycling vertical/horizontal), (5) \textit{other} (stepper, cross trainer, jumping, rowing, basketbatll). 

\subsubsection{Opportunity} The Opportunity dataset \cite{chavarriaga2013opportunity} consists of motion data collected from 16 body-worn sensors sampling at 30 Hz worn by 4 individuals while performing 5 typical daily activities: (1) \textit{standing}, (2) \textit{walking}, (3) \textit{sitting}, (4) \textit{lying}. (5) \textit{other}. However, the dataset contains several repetitive sensor locations. For the purpose of defining distinct domains in our sensor location analysis, we focus on 6 key locations: right wrist, right knee, left upper arm, left shoe, hip, and back.

\subsubsection{RealWorld} The RealWorld dataset \cite{sztyler2016body} contains accelerometer and gyroscope data from 15 participants sampled at 50 Hz. Data is captured from 7 on-body sensors placed at forearm, thigh, head, upper arm, waist, chest, and shin. Each participant performed 8 activities: (1) \textit{climbing stairs down}, (2) \textit{climbing stairs up}, (3) \textit{jumping}, (4) \textit{lying}, (5) \textit{standing}
, (6) \textit{sitting}, (7) \textit{running}, (8) \textit{walking}. 

\subsubsection{HHAR} Heterogeneity Human Activity Recognition dataset \cite{stisen2015smart} covers locomotion activities with accelerometer data collected at the wrist and waist using a wide range of 2 smartwatches (LG G and Samsung Galaxy Gear) and 4 smartphones (LG Nexus 4, Samsung Galaxy S3, Samsung Galaxy S3 mini, and Samsung Galaxy S Plus) respectively. Each device yielded different maximum sampling rates, ranging from 50 Hz to 200 Hz.

% \subsubsection{Lab2Wild} This dataset, we refer to as Lab2Wild in this paper, was part of a study presented by Bhattacharya \textit{et al.} \cite{bhattacharya2022leveraging} that investigated multimodal acoustic and inertial data from a smartwatch capturing a set of 23 complex daily activities such as writing, cooking, cleaning, etc. In this paper, we focus on the accelerometer data only which was sampled at 55 Hz. The dataset contained controlled semi-naturalistic data collected from 15 participants in their own homes as well as \textit{in-the-wild} data from 5 participants. With this dataset, we aimed to study the domain generalization problem under different context environments, in this case different users/homes and controlled to free-living setting.  

\subsection{Data Collection: Juggling Dataset}
\label{sec:juggling}
So far, the public datasets discussed in the previous section cover at least one of the distribution shifts, mainly changes in sensor location, sampling rate, device type, and context environment. An important remaining change that is especially relevant to HAR is change in user behavior. Towards addressing this aspect, we design a data collection study with the goal of capturing a change in user behavior for a certain activity, such as in the case of learning a new skill. Simplifying the problem space, we choose an activity that (1) can be captured with an off-the-shelf smartwatch equipped with accelerometer sensor and (2) can be easily assessed in terms of skill. 

\subsubsection{Study Protocol}
With that in mind, an IRB-approved juggling study was conducted with 9 participants (8 males and 1 female) wearing an Apple Watch on each wrist. The participants were recruited from a local juggling society that includes people from various backgrounds, age (mean 46.1 $\pm$ 19.9), and juggling skill level. The study lasted roughly 30 minutes. We asked participants to wear a watch on each wrist. At the beginning of the data collection study, each participant was asked to complete a survey to collect demographics information as well as a self-reported juggling skill assessment using a 1-5 Likert-Scale question. On average, participant's self-reported skill was 3.4 with lowest score being 2 and highest being 5. 

% \begin{figure}[t]
%     \centering
%     \includegraphics[width = .5\columnwidth]{images/DistributionShift/juggling_stats.png}
%     \caption{Juggling Stats.}
%     \label{fig:juggling_stats}
% \end{figure}

The study was split into three 5-min juggling sessions and four 1-min non-juggling sessions in between. For all participants, the first juggling session entailed juggling 3 balls. Depending on their skill level and the maximum number of balls they are able to juggle, we increased the number of balls juggled in the following sessions. By including sessions with varying numbers of balls, we aimed to capture the progression of skill improvement in juggling, as users gradually increased their ability to juggle more balls over time.  We specifically captured juggling activities with 3, 4, 5, and 7 balls. The number of balls defined in each session for each participant varied based on the participant's skill level. For instance, only one participant was able to juggle 7 balls. For non-juggling activities, participants were asked to perform a set of 4 miscellaneous activities (one minute each) randomly picked from a set of 7 activities including (1) idle activities (standing or sitting), (2) walking, (3) jogging, (4) writing, (5) typing on keyboard, and (6) jumping jacks. 

\begin{figure}[t]
    \centering
    \includegraphics[width = .5\columnwidth]{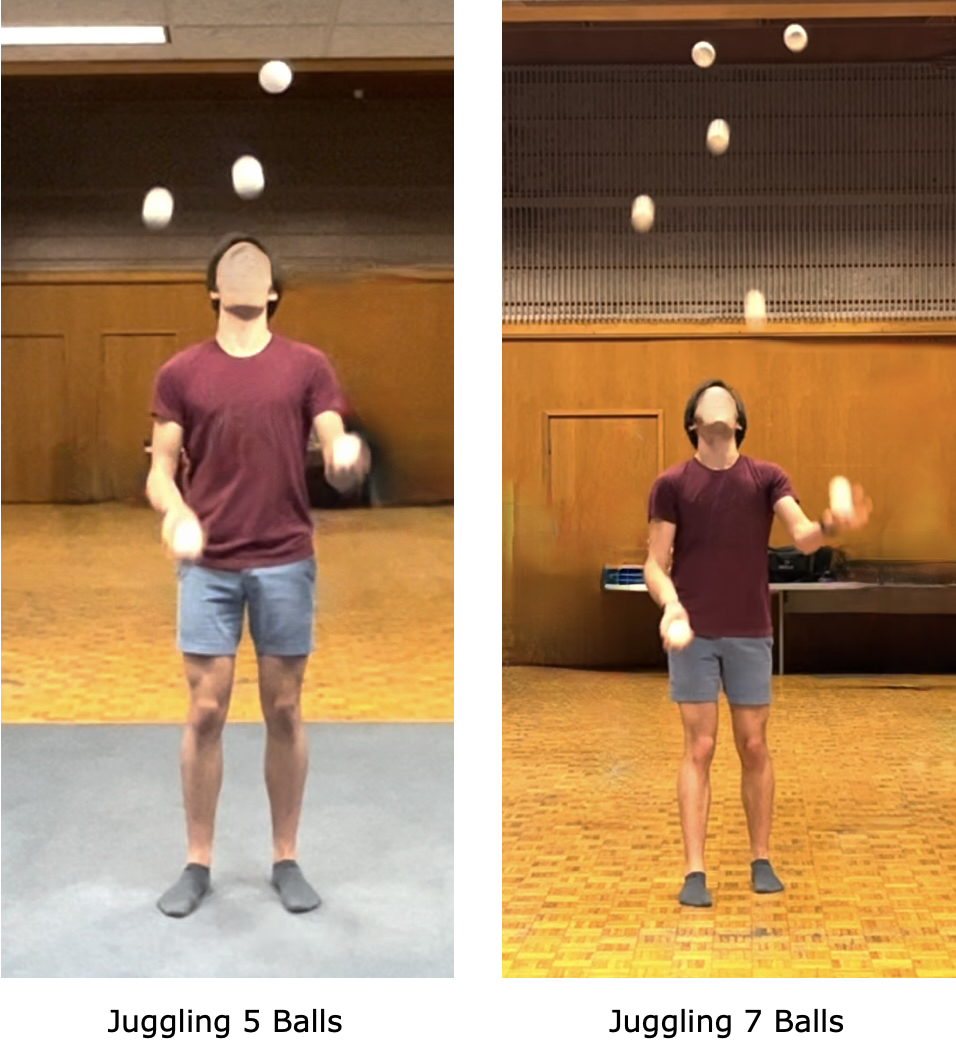}
    \caption{Screenshots from video recorded sessions of a participant juggling 5 and 7 balls.}
    \label{fig:video_juggling}
\end{figure}

Accelerometer data was captured continuously throughout a session at 100 Hz sampling rate using the SensorLog App \cite{thomas2022sensorlog}. The study session was video recorded to aid in data annotation at the end of the study. Figure \ref{fig:video_juggling} depicts example recorded frames of participants juggling 3 and 7 balls. The accelerometer data from both watches was synchronized along with the video recording using the ELAN annotation tool \cite{brugman2004annotating}.

\subsection{Benchmarks}
\label{sec:benchmark}
To investigate the different types of distributions and evaluate the performance of domain generalization algorithms, benchmarks are created for each type of distribution shift based on the characteristics of each dataset. Each setting or change in the sensor data is considered as a domain, and each dataset is preprocessed and prepared to contain a set of $k$ domains with $k > 2$ that follow the corresponding distribution shift. Table \ref{tab:datasets} summarizes the datasets and domains studied for each type of distribution shift. In order to simplify the analysis and have more control over the different types of shifts, we focus on one modality for all datasets, which is the most common and most power efficient sensor--the accelerometer. For all datasets, the 3-axis accelerometer data is first normalized along each axis using standard scaling and then segmented using a fixed sliding window of 100 samples with 50\% overlap. These benchmarks are used for shift quantification and comprehensive performance analysis of the domain generalization algorithms listed in Section \ref{sec:DOG}. We run each experiment 5 times with a random seed and report the average results.  

\subsubsection{Sensor Location} 
In the sensor location analysis, one of the main criteria for dataset selection was the availability of multiple on-body sensor locations capturing synchronized inertial data for common locomotion activities. Based on this criterion, four large datasets were chosen: PAMAP2, DSADS, Opportunity, and RealWorld. Each dataset was divided into separate domains, with each domain representing a specific sensor location. This was done to construct a domain-generalization experiment where the sensor location serves as the domain. Specifically, PAMAP2 was divided into three domains, DSADS into five domains, Opportunity into six domains, and RealWorld into seven domains. 

\subsubsection{Sampling Rate} 
The aim of this domain generalization analysis is to assess the model's ability to generalize to data captured at a different sampling rate from the training data. To ensure that other variables remain consistent, we adopt 3 datasets with high enough sampling rate and simulate the sampling rate changes by downsampling the data. More specifically, we leverage (1) PAMAP2, originally sampled at 100 Hz, and create 3 domains by downsampling to [10Hz, 50Hz]; (2) RealWorld, captured at 50 Hz, and downsample to [10Hz, 25Hz]; and (3) HHAR data captured by LG Nexus4 which was sampled at 200 Hz and downsample it to [10Hz, 50Hz, 100Hz]. Thus, this results in 3,3, and 4 domains for each of the datasets respectively. To reduce data leakage between training and testing data, a stratified k-fold split is applied to the original dataset, with k set to the number of domains in each dataset. Each fold is then preprocessed to correspond to a sampling rate. 

\subsubsection{Device Type}
To explore the domain generalization problem related to sensor heterogeneity across devices, we utilized the HHAR dataset, which consists of locomotion accelerometer data captured from various off-the-shelf smartphones (LG Nexus 4, Samsung Galaxy S3, Samsung Galaxy S3 mini, and Samsung Galaxy S Plus). Each smartphone is equipped with different inertial sensors and samples data at different rates. In this analysis, our main focus was to investigate the impact of sensor biases on data distribution while maintaining other variables constant. To achieve this, we downsampled the data from each device to the lowest sampling rate of 50 Hz. Consequently, data from each device was treated as a distinct domain, resulting in a total of four domains in the HHAR dataset. The objective of this analysis was to examine the generalization capability of models across different devices, which is crucial for the widespread implementation of HAR, given the inherent heterogeneity of devices in real-world scenarios. 

\subsubsection{User Behavior} 
Changes in a user's behavior have been shown to cause distribution shifts. To investigate this aspect, we focused on the scenario of skill improvement, where a user aims to enhance their proficiency in a particular activity, and as such the way they perform the same activity may change over time as their skill level improves. To simulate this setting, we collected a dataset specifically for juggling, capturing juggling activities performed by participants at various skill levels, as explained in Section \ref{sec:juggling}. In this context, we defined domains based on the number of balls being juggled. The underlying assumption was that as users progress in their juggling skills, they would be able to handle an increasing number of balls. To clarify, we divided the juggling activity in the dataset into four domains, with each domain corresponding to a specific number of juggling balls (3, 4, 5, and 7). In order to develop a model that can accurately distinguish juggling from other activities, we applied a stratified four-fold split to the non-juggling activities and included them in each domain dataset. This ensures that each domain dataset contains both juggling and non-juggling activities. A generalizable model should be able to learn the distinguishing features of juggling irrespective of the number of balls.

\section{Quantifying Distribution Shifts}
\label{sec:div_corr}
In general, researchers have defined changes in sensor-based data as a general distribution shift that affects data and, as a result, impacts the performance of a model. However, identifying the cause of the shift and understanding its influence on the data can provide a better understanding of the nature of the distribution shift and guide the selection of the optimal method to address it. Therefore, we aim to first represent and quantify each of the defined causes of distribution shifts in Section \ref{sec:shifts}. Following the characterization by Ye \textit{et al.} \cite{ye2022ood}, we quantify each domain change as a two-dimensional characterization: diversity shift and correlation shift.      
\subsection{Diversity Shift and Correlation Shift}
In plain words, the diversity dimension refers to out-of-distribution samples that are different from the training in-distribution data, while the correlation shift refers to a change in the correlation structure between the input and output variable. With those in mind, a distribution shift can be dominated by one type or a mixture of both, especially in a real-world settings. Following \cite{ye2022ood}, the quantification formula for each of the diversity and correlation shifts between two data distributions $p$ and $q$ is given by, 
$$ D_{div}(p,q) = \frac{1}{2}\int_{\mathcal{S}}|p(z)-q(z)|dz $$
$$ D_{cor}(p,q) = \frac{1}{2}\int_{\mathcal{T}}\sqrt{p(z)\cdot q(z)}\sum_{y\in \mathcal{Y}}|p(y|z)-q(y|z)|dz$$

\noindent where $\mathcal{S}$ and $\mathcal{T}$ represent two sets of features in $\mathcal{Z}$ where $z \in \mathcal{Z}$ possesses the following property, 
$$ p(z)\cdot q(z) = 0 \vee \exists y \in \mathcal{Y} : p(y|z) \neq q(y|z)$$ 
\noindent and $\mathcal{S}$ and $\mathcal{T}$, that are respectively responsible for each diversity and correlation shift, are defined as,
$$ \mathcal{S} := \{z \in \mathcal{Z} | p(z)\cdot q(z) = 0\}$$
$$ \mathcal{T} := \{z \in \mathcal{Z} | p(z)\cdot q(z) \neq 0\}$$

Practically, the metrics can be estimated by training a neural network to discriminate between different domains. Domains can refer to any type of change, whether sensor location, device type, environment etc. as discussed in Section \ref{sec:shifts}. Thus, for example, a neural network is trained to recognize the sensor location the data is captured from. 

\begin{figure}[t]
    \centering
    \includegraphics[width=1.\columnwidth]{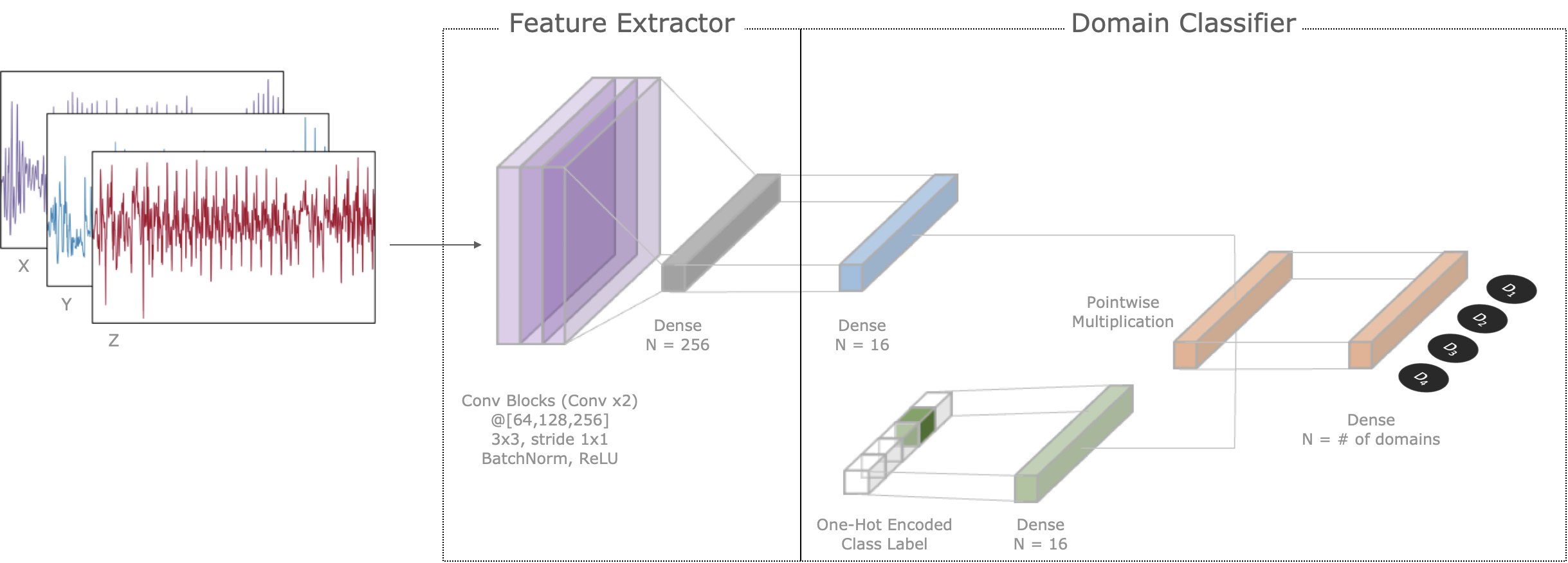}
    \caption{Domain Classifier model overview. The neural network extracts features from the accelerometer data using a Vggish architecture. The domain classifier learns a joint representation of both input and output, which is then used for domain classification.} 
    \label{fig:ood}
\end{figure}

More specifically, to differentiate two datasets $(X_A,Y_A)$ and $(X_B,Y_B)$ with different data distributions $\xi_{A}$ and $\xi_{B}$ respectively, a neural network is trained to distinguish between the two domains. The network is composed of a feature extractor $g: \mathcal{X} \rightarrow \mathcal{F}$ and a domain classifier $h: \mathcal{F} \times \mathcal{Y} \rightarrow [0,1]$, where $\mathcal{F}$ represents a lower dimensional representation of $\mathcal{X}$. This results in a joint distribution over $\mathcal{X} \times \mathcal{Y} \times \mathcal{F}$ for each domain. Figure \ref{fig:ood} illustrates the network framework used to extract features whose joint distribution with $Y$ varies across different domains.

\subsection{Implementation}
Using the benchmark datasets as explained in Section \ref{sec:benchmark}, we apply 5 random stratified splits using domain target label, with 25\% set as test data and 25\% as validation. The network, with a feature dimension m = 16, is optimized using an Adam optimizer with a fixed learning rate of 0.00001 and a batch size of 128 for 100 epochs. The network is trained to optimize a cross-entropy loss. We choose the models maximizing the macro f1-score (in predicting the environments) on validation sets. For estimating the diversity and correlation shifts, an importance sampling size M equal to the smallest class size in the dataset, and we empirically set the thresholds $\epsilon_{div} = 1 \times 10^{-12}$ and $\epsilon_{cor} = 5 \times 10^{-4}$. We use Gaussian kernels for all the KDEs.

\section{Domain Generalization}
From a machine learning perspective, the challenge of data diversity in HAR models can be formulated as a \textit{domain shift} problem. To tackle the aforementioned challenges, it is crucial to assess the generalizability of a model on novel data. An effective approach frequently employed to address domain shift issues is transfer learning or domain adaptation \cite{gjoreski2019cross,10.1145/2971763.2971764,wang2018deep,zhao2020local}, which involves utilizing knowledge acquired from other available domains during the training phase, and then fine-tuning the model on a new target domain to minimize the distribution gap and improve overall performance. Nevertheless, a primary assumption for such techniques is the availability or accessibility of the target data, which is often unfeasible in practical applications. This is particularly true in the realm of sensor-based activity data, where it is not practical to amass sensor data for all potential distribution changes. Thus, domain generalization is a research topic that focuses on the problem of building a model that generalizes well to unseen data without access to the new data during training. 

As noted earlier, the challenge of generalization in HAR can arise from various factors, such as changes in device type, sensor heterogeneity in terms of sampling rate or sensor placement, and activity patterns, among others (see Section \ref{sec:shifts}). While these challenges have been recognized in previous research, it is still unclear to the scientific community which of these factors specifically makes domain generalization in HAR particularly difficult. In this chapter, we take the initial step in conducting a methodical evaluation of the generalizability of HAR models across these distribution changes. 
% We urge researchers to include this critical step in their model analysis pipeline to further advance the understanding of HAR model performance in real-world scenarios.

% To perform this analysis, we first investigate the different causes of distribution shifts, focusing on sensor-related changes (e.g. device type, sensor placement and sampling rate), user-related changes (e.g. skill improvement), and environment-related changed (e.g. controlled vs. in-the-wild). Taking inspiration from the work of Ye \textit{et al.} \cite{ye2022ood}, we quantify each distribution shift using a two-dimensional spectrum, which includes measurements of diversity shift and correlation shift. Second, 
There have been significant advancements in domain generalization methods within the machine learning community, particularly in the fields of computer vision and natural language processing \cite{zhou2022domain}. However, the problem of domain generalization in HAR has received limited attention. To the best of our knowledge, our research is the first to investigate domain generalization in the context of specific distribution shifts relevant to sensor-based HAR, with a focus on understanding the challenges each shift poses to model generalizability. In particular, our main research question is: \textit{how effective are domain generalization algorithms in realistic HAR scenarios?} To answer this question, we developed a uniform benchmark platform and evaluated up to 28 domain generalization methods on 4 types of distribution shifts.

\subsection{Problem Formulation}

Let $\mathcal{D} = \{D_1, D_2, ..., D_k\}$ be a set of $k$ domains, where each domain $D_i$ is a joint distribution over feature space $\mathcal{X}$ and label space $\mathcal{Y}$. In the HAR context, a domain corresponds to a particular setting or condition in which the sensor data was collected (e.g., different device types, sensor locations, or environments). The goal of domain generalization is to learn a model $f_\theta : \mathcal{X} \rightarrow \mathcal{Y}$ that performs well on a new, unseen domain $D_{k+1}$, without having access to any labeled data from that domain during training.

Formally, given the set of $k$ domains $\mathcal{D}$, a domain generalization algorithm seeks to find a model $\theta$ that minimizes the expected risk over the joint distribution of all the domains:

$$
\min_{\theta} \frac{1}{k} \sum_{i=1}^{k} \mathbb{E}_{(x,y) \sim D_i} [ \ell(y, f_\theta(x))] 
$$

where $\ell(y, f_\theta(x))$ is a loss function that measures the prediction error of the model $f_\theta$ on a given input $(x,y)$, and $\mathbb{E}_{(x,y) \sim D_i}$ is the expected value taken over the joint distribution $D_i$. The idea is to learn a model that generalizes well to the unseen domain $D_{k+1}$ by minimizing the differences among the learned models from each domain in $\mathcal{D}$. 

In practice, this is achieved by exploiting the commonalities among the domains and learning features that are invariant to the distribution shifts, such that the model can generalize to new domains. 

\subsection{Algorithms}
\label{sec:DOG}
Leveraging DomainBed \cite{gulrajani2020search}, we evaluated 28 baseline techniques to cover the major approaches of domain generalization, including (1) data manipulation, (2) representation learning, and (3) learning strategy. 
\setlist{nolistsep}
\begin{enumerate}[noitemsep]
    \item \textbf{ERM} (Empirical Risk Minimization) \cite{vapnik1999overview} is a standard approach to training machine learning models on a single dataset. In domain generalization, ERM is modified to train models on multiple source domains without access to target domain data.
    \item \textbf{Fish} (First-Order Gradient Matching) \cite{shi2021gradient} is a first-order algorithm that solves an inter-domain gradient matching objective between gradients of different domains.
    \item \textbf{IRM} (Invariant Risk Minimization) \cite{arjovsky2019invariant} is a regularization-based approach that encourages models to be invariant to domain shifts. It seeks a representation that remains invariant across multiple domains.
    \item \textbf{GroupDRO} (Group Distributionally Robust Optimization) \cite{sagawa2019distributionally} is a robust optimization-based approach that addresses the domain shift problem by minimizing the worst-case empirical risk over a set of data groups. 
    \item \textbf{Mixup} (ERM-Mixup) \cite{yan2020improve} is a data manipulation and augmentation technique that applies a weighted linear interpolation between pairs of examples from different domains.   
    \item \textbf{MLDG} (Meta-Learning for Domain Generalization) \cite{li2018learning} leverages a meta-learning strategy to meta-learn how to generalize to various domains. 
    \item \textbf{CORAL} (Correlation Alignment) \cite{sun2016deep} aligns the second-order statistics of feature distributions. 
    \item \textbf{MMD} (Maximum Mean Discrepancy) \cite{li2018domain} is a kernel-based measure of distribution distance. It is used to minimize the difference between the feature distributions of the source and target domains.
    \item \textbf{DANN} (Domain-Adversarial Neural Network) \cite{ganin2016domain} trains a feature extractor to be domain-invariant by adding a domain classification head as a discriminator and optimizing the adversarial loss. 
    \item \textbf{CDANN} (Conditional Domain-Adversarial Neural Network) \cite{li2018deep} extends DANN by adding a conditional feature extractor that takes into account the source and target domain labels.
    \item \textbf{MTL} (Marginal Transfer Learning) \cite{blanchard2021domain} is a transfer learning-based approach that learns shared representations across multiple domains and a domain-specific classifier for each domain.
    \item \textbf{SagNet} (Style Agnostic Network) \cite{nam2021reducing} is a domain generalization approach that learns to ignore style-specific information by maximizing the distance between style-specific and content-specific feature distributions.
    \item \textbf{ARM} (Adaptive Risk Minimization) \cite{zhang2021adaptive} is a risk minimization-based approach that reduces the risk of making a wrong prediction on the target domain by minimizing the empirical risk with an adaptive penalty term.
    \item \textbf{VREx} (Variance Risk Extrapolation) \cite{krueger2021out} is a regularization-based approach that extrapolates the variance of the model's predictions from the source domains to the target domain to reduce the risk of overfitting.
    \item \textbf{RSC} (Representation Self-Challenging) \cite{huang2020self} is a self-supervised learning approach that trains a model to predict the representation of a transformed input and then minimizes the distance between the predicted and actual representations to learn more robust features.
    \item \textbf{SD} (Spectral Decoupling) \cite{pezeshki2021gradient} is a data transformation-based approach that decouples the spectral components of the data to improve generalization performance. 
    \item \textbf{ANDMask} (Learning Explanations that are Hard to Vary) \cite{parascandolo2020learning} an explanation method that produces explanations that are robust to small perturbations in the input. It does this by masking out input features that are not important for the model's output.
    \item \textbf{IGA} (Inter Gradient Alignment) \cite{koyama2020invariance} proposes a method for improving out-of-distribution generalization by finding the maximal invariant predictor of the output.
    \item \textbf{SelfReg} (Self-supervised Contrastive Regularization) \cite{kim2021selfreg} trains a network to distinguish between pairs of augmented versions of the same sample data, and pairs of data from different classes. 
    \item \textbf{SANDMask} (Smoothed-AND Mask) \cite{shahtalebi2021sand} proposes an explanation method that produces smoothed-AND masks, which are more robust to adversarial attacks than traditional masks. The smoothed-AND mask is computed by taking the minimum of the activation map and a Gaussian kernel.
    \item \textbf{TRM} (Transfer Risk Minimization) \cite{xu2021learning} proposes a method for learning representations that support the transfer of predictors across different domains. 
    \item \textbf{IB\_ERM} (Information Bottleneck-ERM) \cite{ahuja2021invariance} combines the invariance principle with the information bottleneck principle, which encourages the model to learn representations that are both invariant to changes in the input distribution and informative for the task.
    \item \textbf{IB\_IRM} (Information Bottleneck-IRM) \cite{ahuja2021invariance} is similar to IB-ERM but uses a different training objective based on IRM instead of ERM.
    \item \textbf{CAD} (Contrastive Adversarial Domain) \cite{ruan2022optimal} proposes a method for learning representations that are optimal for handling covariate shift, by learning a contrastive adversarial domain adaptation model.
    \item \textbf{CondCAD}(Conditional Contrastive Adversarial Domain) \cite{ruan2022optimal} proposes a conditional variation of CAD.
    \item \textbf{Transfer} \cite{zhang2021quantifying} quantifies and analyzes the transferability of features learned by using a similarity measure of features between domains. It then augments the training data with domain-specific transformations for domain generalization.
    \item \textbf{CausIRL} (Invariant Representation Learning) \cite{chevalley2022invariant} uses the invariant representation Learning to learn causal mechanisms that are invariant across different environments. It leverages domain adaptation techniques such as CORAL or MMD to match the distributions of features across environments and trains a model that captures the underlying causal structure of the data.
    \item \textbf{EQRM} (Empirical Quantile Risk Minimization) \cite{eastwood2022probable}    aims to improve the robustness of machine learning models to outliers by minimizing the empirical quantile risk of the model. It proposes a new loss function that is a weighted sum of the quantile regression losses at different quantiles of the target distribution,  where the weights are determined by the expected loss under the target distribution. The proposed method is shown to be more robust to outliers than traditional mean-squared error loss.   
\end{enumerate}

\subsection{Evaluation}

To ensure fair and reliable performance evaluation of the benchmarks and algorithms, we developed a benchmark platform inspired by DomainBed (Gulrajani et al., 2020) that incorporates all the algorithms and prepares the datasets as described in Section \ref{sec:benchmark}. For model selection, we employ a technique called "leave-one-domain-out validation." This approach involves iteratively splitting the \textit{k} domains in a dataset, where at each iteration, one domain is held out as the test domain, while the remaining \textit{k-1} domains are used for training. Additionally, within the \textit{k-1} training domains, a portion is set aside as a validation set for hyperparameter tuning. The objective is to evaluate the model's performance not only on the training domains but also on unseen test domains. Therefore, the model that achieves the highest combined performance on both the validation and test domains is selected and reported as the best domain generalization model. This methodology ensures that the chosen model demonstrates robust generalization across different domains. 

In our evaluation, we report the accuracy as the evaluation metric for each domain in the dataset. We calculate the \textit{in\_accuracy} for each domain in the validation set and the \textit{out\_accuracy} for the test domain. These accuracy values provide insights into how well the model performs on individual train and test domains. To compare the algorithms, we compute the average performance across both the training and test domains. Additionally, we calculate a ranking score for each algorithm relative to the ERM approach, indicating whether the performance is lower than ($-1$), within [-1,1] difference ($0$), or higher ($+1$) than the baseline ERM method.

\section{Results}
\begin{table*}[t]
    \caption{Estimation of diversity and correlation shift for each dataset and type of distribution shift.}
    \centering
    \small
    \begin{tabular}{@{}l|l|c|c@{}}
    \toprule
        \textbf{Distribution Change} & \textbf{Dataset} & \textbf{Div. Shift} & \textbf{Corr. Shift} \\
        \hline
        \multirow{3}{*}{Sensor Location} & PAMAP2 & 0.058 $\pm$ 0.04 &  0.00 $\pm$ 0.00 \\
        & DSADS & 0.997 $\pm$0.14 &  0.00 $\pm$ 0.00 \\
        & Opportunity & 0.768 $\pm$ 0.2 & 0.00 $\pm$ 0.00 \\
        % & RealWorld & & \\
        \hline
        \multirow{3}{*}{Sampling Rate} & PAMAP2 & 0.586 $\pm$ 0.30 & 0.00 $\pm$ 0.00 \\
        & RealWorld & 0.291 $\pm$ 0.21 &  0.00 $\pm$ 0.00 \\
        & HHAR Nexus4 &  0.632 $\pm$ 0.21 &  0.00 $\pm$ 0.00 \\
        \hline
        Device Type & HHAR & 0.857 $\pm$ 0.04 &  0.00 $\pm$ 0.00 \\
        % \hline
        % Context Environment & Lab2Wild & & \\
        \hline
        User Behavior & Juggling &0.733 $\pm$ 0.17 &  0.00 $\pm$ 0.00 \\
        \bottomrule
    \end{tabular}
    \label{tab:div_corr}
\end{table*}

\subsection{Quantifying Distribution Shifts}
Based on the quantification formulas for diversity and correlation shifts, we computed the two-dimensional shift for each shift type and dataset. The results are summarized in Table \ref{tab:div_corr}. 

From the analysis, we observe that all types of shifts and datasets are predominantly characterized by diversity shift, while exhibiting zero correlation shift.  This implies that, for all shift types (sensor location, sampling rate, device type, and user behavior) and datasets in question, the distribution shift in the data is embodied by novel features that are not shared across domains and have no correlation with the activity classes. Thus, as demonstrated in the domain generalization analysis in the following section, creating a model that can effectively generalize to various distribution shifts is a challenging task. 

\begin{table*}[t]
    \caption{Average performance (\textit{in\_accuracy} / \textit{out\_accuracy}) of ERM and domain generalization algorithms on different datasets dominated by different shift types.} 
    \centering
    \tiny
    \begin{tabularx}{\textwidth}{@{}X|YYY|YYY|Y|Y@{}}
         & \multicolumn{3}{c|}{\textbf{Sensor Location}} & \multicolumn{3}{c|}{\textbf{Sampling Rate}} & \textbf{Device Type} & \textbf{User Behavior}\\
        \cmidrule{2-9}
        & PAMAP2 & DSADS & Opportunity & PAMAP2 & RealWorld & HHAR-Nexus4 & HHAR & Juggling \\
        \hline
        ERM & 0.83 / 0.34 & 0.80 / 0.58 & 0.52 / 0.37 & 0.98 / 0.88 & 0.96 / 0.75 & 0.80 / 0.58 & 0.80 / 0.72 & 0.97 / 0.89 \\
        \hdashline
        Fish & 0.74 / 0.34 & 0.74 / 0.54 & 0.46 / 0.38 & 0.97 / 0.88 & 0.95 / 0.74 &  0.74 / 0.57 & \textbf{\underline{0.84 / 0.76}} & 0.97 / 0.90 \\
        IRM & 0.82 / 0.35 & 0.77 / 0.57 & 0.50 / 0.37 & 0.98 / 0.88 & 0.95 / 0.77 & 0.79 / 0.57 & 0.78 / 0.71 & 0.97 / 0.88\\
        GroupDRO & 0.69 / 0.35 & 0.68 / 0.53 & 0.44 / 0.37 & 0.98 / 0.88 & 0.94 / 0.73 & 0.72 / 0.58 & 0.81 / 0.74 & 0.97 / 0.90\\
        Mixup & 0.60 / 0.33 & 0.65 / 0.50 & 0.42 / 0.35 & 0.97 / 0.86 & 0.92 / 0.74 & 0.72 / 0.56 & 0.77 / 0.70 & 0.97 / 0.89\\
        MLDG & 0.14 / 0.14 & 0.28 / 0.27 & 0.27 / 0.27 & 0.15 / 0.15  & 0.06 / 0.06 &  0.30 / 0.30 & 0.30 / 0.30 & 0.57 / 0.56\\
        CORAL & 0.71 / 0.32 & 0.68 / 0.54 & 0.44 / 0.36 & 0.97 / 0.88 & 0.95 / 0.74 & 0.72 / 0.57 & 0.80 / 0.73 & 0.97 / 0.90\\
        MMD & 0.72 / 0.33 & 0.69 / 0.54 & 0.45 / 0.38 & 0.97 / 0.87 & 0.95 / 0.73 &  0.72 / 0.56 & 0.81 / 0.72 & 0.97 / 0.89\\
        DANN & 0.77 / 0.34 & 0.72 / 0.53 & 0.48 / 0.38 & 0.96 / 0.88 & 0.94 / 0.75 & 0.75 / 0.57 & 0.75 / 0.69 & 0.96 / 0.89\\
        CDANN & 0.78 / 0.33 & 0.73 / 0.52 & 0.49 / 0.40 & 0.96 / 0.88 & 0.93 / 0.77 & 0.74 / 0.57 & 0.75 / 0.69 & 0.96 / 0.89\\
        MTL & 0.70 / 0.32 & 0.68 / 0.52 & 0.43 / 0.36 & 0.98 / 0.87 & 0.94 / 0.72 & 0.71 / 0.58 & 0.79 / 0.73 & \textbf{\underline{0.97 / 0.91}}\\
        SagNet & 0.83 / 0.33 & 0.78 / 0.57 & 0.50 / 0.36 & 0.98 / 0.89 & 0.96 / 0.76 & \textbf{\underline{0.81 / 0.60}} & 0.81 / 0.74 & 0.97 / 0.90\\
        ARM & 0.84 / 0.34 & 0.81 / 0.53 & 0.52 / 0.35 & 0.98 / 0.89 & 0.97 / 0.77 & 0.80 / 0.55 & 0.81 / 0.72 & 0.97 / 0.89\\
        VREx & 0.83 / 0.34 & 0.78 / 0.58 & 0.51 / 0.35 & 0.98 / 0.87 & 0.96 / 0.74 & 0.79 / 0.58 & 0.80 / 0.73 & 0.97 / 0.89\\
        RSC & 0.83 / 0.35 & 0.78 / 0.58 & 0.50 / 0.37 & \textbf{\underline{0.98 / 0.89}} & 0.96 / 0.76 & 0.78 / 0.60 & 0.79 / 0.72 & 0.98 / 0.90\\
        SD & 0.83 / 0.34 & 0.79 / 0.57 & 0.51 / 0.36 & 0.98 / 0.88 & 0.97 / 0.77 & 0.82 / 0.58 & 0.80 / 0.73 & 0.97 / 0.89\\
        ANDMask & 0.69 / 0.37 & 0.66 / 0.51 & 0.44 / 0.39 & 0.97 / 0.87 & 0.94 / 0.73 & 0.71 / 0.57 & 0.79 / 0.73 & 0.97 / 0.90\\
        SANDMask & 0.22 / 0.24 & 0.09 / 0.09 & 0.07 / 0.08 & 0.16 / 0.17 & 0.13 / 0.13 & 0.09 / 0.10 & 0.09 / 0.09 & 0.14 / 0.09\\
        IGA & 0.32 / 0.29 & 0.18 / 0.17 & 0.12 / 0.12 & 0.43 / 0.42 & 0.27 / 0.26 & 0.25 / 0.26 & 0.18 / 0.18 & 0.53 / 0.57\\
        SelfReg & 0.83 / 0.33 & 0.79 / 0.58 & 0.50 / 0.36 & 0.98 / 0.89 & \textbf{\underline{0.97 / 0.76}} & 0.81 / 0.59 & 0.81 / 0.73 & 0.97 / 0.89\\
        TRM & \textbf{\underline{0.83 / 0.36}} & 0.80 / 0.57 & \textbf{\underline{0.52 / 0.38}} & 0.98 / 0.87 & 0.96 / 0.77 & 0.80 / 0.58 & 0.80 / 0.72 & 0.97 / 0.89\\
        IB (ERM) & 0.84 / 0.33 & \textbf{\underline{0.80 / 0.59}} & 0.51 / 0.36 &0.98 / 0.88  & 0.96 / 0.75 & 0.81 / 0.59 & 0.80 / 0.73 & 0.97 / 0.90\\
        IR (IRM)  & 0.83 / 0.32 & 0.77 / 0.58 & 0.49 / 0.35 & 0.98 / 0.88 & 0.96 / 0.74 & 0.79 / 0.56 & 0.78 / 0.71 & 0.97 / 0.89\\
        CAD & 0.49 / 0.38 & 0.46 / 0.37 & 0.41 / 0.34 & 0.71 / 0.68 & 0.49 / 0.45 & 0.50 / 0.44 & 0.48 / 0.47 & 0.73 / 0.72\\
        CondCAD & 0.53 / 0.34 &  0.47 / 0.38 & 0.40 / 0.36 & 0.73 / 0.68 & 0.61 / 0.51 & 0.47 / 0.42 & 0.54 / 0.52 & 0.73 / 0.73\\
        Transfer & 0.74 / 0.35 & 0.65 / 0.52 & 0.42 / 0.33 & 0.97 / 0.90 & 0.95 / 0.76 & 0.70 / 0.58 & 0.73 / 0.70 & 0.95 / 0.88\\
        % CausIRL (CORAL) & 0.71 / 0.34 & 0.68 / 0.52 & 0.44 / 0.37 & 0.97 / 0.87 & 0.95 / 0.72 & 0.71 / 0.55 & 0.80 / 0.74 & 0.97 / 0.90\\
        CausIRL & 0.69 / 0.37 & 0.68 / 0.53 & 0.44 / 0.38 & 0.97 / 0.88 & 0.94 / 0.75 & 0.71 / 0.57 & 0.80 / 0.73 & 0.97 / 0.90\\
        EQRM & 0.69 / 0.36 & 0.68 / 0.53 & 0.44 / 0.37 & 0.97 / 0.88 & 0.95 / 0.73 & 0.72 / 0.58 & 0.80 / 0.73 & 0.97 / 0.89\\
        \bottomrule
    \end{tabularx}
    \label{tab:dg}
\end{table*}

\begin{table*}[t]
    \caption{Ranking score of generalization algorithms w.r.t. ERM on different datasets dominated by different shift types. Values marked with * indicate that the difference in performance across 5 runs compared to ERM baseline are statistically significant with p-value $< 0.05$.} 
    \centering
    \tiny
    \begin{tabularx}{\textwidth}{@{}X|YYY|YYY|Y|Y@{}}
         & \multicolumn{3}{c|}{\textbf{Sensor Location}} & \multicolumn{3}{c|}{\textbf{Sampling Rate}} & \textbf{Device Type} & \textbf{User Behavior}\\
        \cmidrule{2-9}
        & PAMAP2 & DSADS & Opportunity & PAMAP2 & RealWorld & HHAR-Nexus4 & HHAR & Juggling \\
        \hline
        % ERM & - & - & - & - & - & - & - & - \\
        % \hdashline
        Fish & $-1$ * & $-1$ * & $-1$ * & 0 & $-1$ & $-1$ * & $+1$ * & 0 \\
        IRM & 0 & $-1$ * & $-1$ & 0 & 0 & $-1$ & $-1$ * & $-1$\\
        GroupDRO & $-1$ * & $-1$ * & $-1$ * & 0 & $-1$ * & $-1$ * & $+1$ & 0\\
        Mixup & $-1$ * & $-1$ * & $-1$ * & $-1$ & $-1$ * & $-1$ * & $-1$ * & 0\\
        MLDG & $-1$ * & $-1$ * & $-1$ * & $-1$ * & $-1$ * & $-1$ * & $-1$ * & $-1$ *\\
        CORAL & $-1$ * & $-1$ * & $-1$ * & 0 & $-1$ & $-1$ * & 0 & 0\\
        MMD & $-1$ * & $-1$ * & $-1$ * & $-1$ & $-1$ & $-1$ * & $+1$ & 0\\
        DANN & $-1$ * & $-1$ * & $-1$ * & $-1$ & $-1$ & $-1$ * & $-1$ * & $-1$ *\\
        CDANN & $-1$ * & $-1$ * & 0 & $-1$ & $-1$ & $-1$ * & $-1$ * & $-1$ *\\
        MTL & $-1$ * & $-1$ * & $-1$ * & $-1$ & $-1$ * & $-1$ * & 0 & $+1$ *\\
        SagNet & $-1$ & $-1$ & $-1$ * & $+1$ & $+1$ & $+1$ * & $+1$ * & 0\\
        ARM & $+1$ & $-1$ * & $-1$ & $+1$ & $+1$ & $-1$ & 0 & 0\\
        VREx & 0 & $-1$ & $-1$ * & 0 & 0 & $-1$ & 0 & 0\\
        RSC & 0 & $-1$ & $-1$ & $+1$* & 0 & 0 & $-1$ & $+1$\\
        SD & 0 & $-1$ & $-1$ & 0 & $+1$ & $+1$ & 0 & 0\\
        ANDMask & $-1$ * & $-1$ * & $-1$ * & $-1$ & $-1$ * & $-1$ * & 0 & 0\\
        SANDMask & $-1$ * & $-1$ * & $-1$ * & $-1$ * & $-1$ * & $-1$ * & $-1$ * & $-1$ *\\
        IGA & $-1$ * & $-1$ * & $-1$ * & $-1$ * & $-1$ * & $-1$ * & $-1$ * & $-1$ *\\
        SelfReg & 0 & $-1$ & $-1$ * & $+1$ & $+1$ & $+1$ & $+1$ & 0\\
        TRM & $+1$ & $-1$ & 0 & 0 & $+1$ & 0 & 0 & 0\\
        IB (ERM) & 0 & 0 & $-1$ & 0 & 0 & $+1$ & 0 & 0\\
        IB (IRM)  & $-1$ & $-1$ * & $-1$ * & 0 & $-1$ & $-1$ & $-1$ * & $-1$\\
        CAD & $-1$ * & $-1$ * & $-1$ * & $-1$ * & $-1$ & $-1$ * & $-1$ * & $-1$ *\\
        CondCAD & $-1$ * &  $-1$ * & $-1$ * & $-1$ * & $-1$ * & $-1$ * & $-1$ * & $-1$ *\\
        Transfer & $-1$ * & $-1$ * & $-1$ * & $+1$ * & 0 & $-1$ * & $-1$ * & $-1$ *\\
        % CausIRL (CORAL) & 0.71 / 0.34 & 0.68 / 0.52 & 0.44 / 0.37 & 0.97 / 0.87 & 0.95 / 0.72 & 0.71 / 0.55 & 0.80 / 0.74 & 0.97 / 0.90\\
        CausIRL & $-1$ * & $-1$ * & $-1$ * & 0 & $-1$ & $-1$ * & 0 & 0\\
        EQRM & $-1$ * & $-1$ * & $-1$ * & 0 & $-1$ & $-1$ * & 0 & 0\\
        \bottomrule
    \end{tabularx}
    \label{tab:rank_score}
\end{table*}

\subsection{Domain Generalization Analysis}

One of our key research goals is to explore the generalizability of human activity recognition models across different types of distribution shifts. Using the benchmark experiments discussed in Section \ref{sec:benchmark}, we evaluate the performance of several domain generalization algorithms (Section \ref{sec:DOG}).  

% The benchmark results are shown in Tables \ref{tab:dg} and \ref{tab:rank_score}.
Table \ref{tab:dg} reports the average \textit{in\_accuracy} and \textit{out\_accuracy} over 5 runs for each dataset/shift-algorithm pair. We, further, report a ranking score for each algorithm in Table \ref{tab:rank_score}. We underline that the ranking score does not indicate whether an algorithm is definitely better or worse than the other algorithms, but instead reflects a relative degree of robustness against a distribution shift. From Tables \ref{tab:dg} and \ref{tab:rank_score}, we observe that none of the domain generalization algorithms achieves consistently better performance than ERM on all types of distribution shifts and datasets. 

For sensor location shift, almost all algorithms achieve lower performance compared to ERM on all 3 datasets. Similarly, for sampling rate shift, we observe an improvement in performance for SelfReg across all datasets, however the results are not statistically significant. For device type, Fish achieves a statistically significant improvement in performance compared to ERM, while MTL improves performance for user behavior shift.

\section{Discussion}

This work provides a unified benchmark framework to evaluate several domain generalization methods on multiple distribution shift scenarios. Our experiments unveil various key insights on the challenges of domain generalization. 

\subsection{Distribution Gap and Challenges of Domain Generalization}

Investigating various types of distribution shifts, we observe a large performance gap as a result of the distribution shift caused by changes in either the wearable sensing device, sensor characteristics, or user's activity pattern. These changes can induce feature distribution bias into a model, making it challenging to generalize well to diverse datasets and data collection setups. 

The performance results of ERM presented in Table \ref{tab:dg} reveal that distinct shift types and datasets exhibit varying performance gaps between the \textit{in\_{accuracy}} and \textit{out\_{accuracy}}. Specifically, changes in sensor location induce a feature distribution shift, resulting in an approximately 0.2 performance gap for DSADS and Opportunity, and even a substantial drop in performance ($\sim$0.5) for PAMAP2. None of the domain generalization algorithms demonstrate an an improved performance on the \textit{out-of-distribution} target data, underscoring the considerable challenge of domain generalization. On the other hand, a distribution shift caused by changes in sampling rate yield a performance gap of $\sim$0.2 in ERM for RealWorld and HHAR-Nexus4, and $\sim$ 0.1 for PAMAP2. The smaller performance gap for PAMAP2 may be attributed to the smaller number of activities being learned, leading to a more simplified problem. In this scenario, applying domain generalization leads to a statistically significant slight improvement in performance for PAMAP2 and HHAR-Nexus4 datasets. A small performance gap ($\sim$0.08) is also observed for shifts in device type and user behavior, with certain domain generalization methods effectively mitigating this performance gap. This suggests that challenges in domain generalization are more pronounced with shifts in sensor location and sampling rate compared to temporal user behavior changes and variations in device type.

% \subsection{Challenges of Domain Generalization}

\begin{table*}[t]
    \caption{Performance (average \textit{target\_{accuracy}} $\pm$ standard deviation) of domain adaptation algorithms on different datasets and cross-domain scenarios dominated by different shift types.} 
    \centering
    \tiny
    \begin{tabularx}{\textwidth}{@{}X|YYY|YYY|Y|Y@{}}
         & \multicolumn{3}{c|}{\textbf{Sensor Location}} & \multicolumn{3}{c|}{\textbf{Sampling Rate}} & \textbf{Device Type} & \textbf{User Behavior}\\
        \cmidrule{2-9}
        & PAMAP2 & DSADS & Opportunity & PAMAP2 & RealWorld & HHAR-Nexus4 & HHAR & Juggling\\
        \hline
        Target Only & $0.91 \pm 0.033$ & $0.97 \pm 0.001$ & $0.71 \pm 0.008$ & $0.98 \pm 0.009$ & $0.99 \pm 0.001$ & $0.94 \pm 0.004$ & $0.96 \pm 0.005$ & $0.99 \pm 0.001$\\
        \hdashline
        Deep Coral & $0.44 \pm 0.022$ & $0.48 \pm 0.006$ & $0.30 \pm 0.010$ & $0.80 \pm 0.040$ & $0.58 \pm 0.023$ &  $0.39 \pm 0.005$ & $0.71 \pm 0.006$ & $0.72 \pm 0.010$\\
        SASA & $0.40 \pm 0.028$ & $0.45 \pm 0.018$ & $0.38 \pm 0.015$ & $0.84 \pm 0.007$ & $0.73 \pm 0.009$ & $0.47 \pm 0.014$ & $0.76 \pm 0.006$ & $0.69 \pm 0.006$\\
        CoDATS & $0.48 \pm 0.028$ & \underline{$\bm{0.55 \pm 0.026}$} & $0.36 \pm 0.016$ & $0.83 \pm 0.003$ & $0.68 \pm 0.018$ & $0.48 \pm 0.016$ & \underline{$\bm{0.81 \pm 0.005}$} & $0.69 \pm 0.016$\\
        DSAN & $0.36 \pm 0.055$ & $0.37 \pm 0.035$ & $0.39 \pm 0.036$ & $0.79 \pm 0.025$ & \underline{$\bm{0.75 \pm 0.016}$} & \underline{$\bm{0.55 \pm 0.015}$} & $0.80 \pm 0.007$ & $0.65 \pm 0.027$ \\
        AdvSKM & $0.40 \pm 0.032$ & $0.47 \pm 0..019$ & $0.29 \pm 0.018$ & $0.81 \pm 0.025$  & $0.59 \pm 0.019$ &  $0.39 \pm 0.008$ & $0.71 \pm 0.008$ & $0.71 \pm 0.016$\\
        DIRT & $0.37 \pm 0.021$ & $0.40 \pm 0.020$ & $0.37 \pm 0.028$ & \underline{$\bm{0.87 \pm 0.025}$} & $0.65 \pm 0.039$ & $0.42 \pm 0.034$ & $0.78 \pm 0.007$ & $0.65 \pm 0.056$ \\
        CDAN & $0.44 \pm 0.044$ & $0.49 \pm 0.036$ & \underline{$\bm{0.43 \pm 0.033}$} & \underline{$\bm{0.87 \pm 0.044}$} & $0.72 \pm 0.032$ &  $0.49 \pm 0.032$ & \underline{$\bm{0.81 \pm 0.006}$} & $0.71 \pm 0.009$\\
        MMDA & $0.44 \pm 0.051$ & $0.47 \pm 0.024$ & $0.30 \pm 0.10$ & $0.76 \pm 0.035$ & $0.59 \pm 0.013$ & $0.41 \pm 0.014$ & $0.71 \pm 0.011$ & \underline{$\bm{0.78 \pm 0.022}$}\\
        DANN & \underline{$\bm{0.49 \pm 0.035}$} & $0.53 \pm 0.025$ & $0.38 \pm 0.016$ & $0.86 \pm 0.018$ & $0.69 \pm 0.031$ & $0.46 \pm 0.031$ & $0.80 \pm 0.010$ & $0.69 \pm 0.019$\\
        DDC & $0.43 \pm 0.041$ & $0.46 \pm 0.016$ & $0.31 \pm 0.017$ & $0.80 \pm 0.027$ & $0.58 \pm 0.023$ & $0.39 \pm 0.013$ & $0.71 \pm 0.005$ & $0.72 \pm 0.014$\\
        CoTMix & $0.36 \pm 0.047$ & $0.37 \pm 0.034$ & $0.17 \pm 0.002$ & $0.64 \pm 0.040$ & $0.57 \pm 0.025$ & $0.18 \pm 0.024$ & $0.27 \pm 0.095$ & $0.63 \pm 0.038$\\
        \hdashline
        No Adapt & $0.35 \pm 0.016$ & $0.44 \pm 0.025$ & $0.24 \pm 0.008$ & $0.76 \pm 0.029$ & $0.53 \pm 0.025$ & $0.36 \pm 0.013$ & $0.67 \pm 0.013$ & $0.74 \pm 0.059$\\
        \bottomrule
    \end{tabularx}
    \label{tab:da}
\end{table*}

\subsection{Domain Adaptation and Model Personalization}
Considering the challenges inherent in domain generalization, there is a compelling argument to transition towards domain adaptation, wherein the model adapts to the target domain given a limited subset of target data. In real-world scenarios, data is susceptible to endless distribution shifts that prove challenging for generalization. Additionally, it is practically impossible to account for all distribution shifts and generalize to all of them. Consequently, a more viable approach involves enabling a model to adapt and personalize to new domains, be it a new device, sensor, or individual. 

With this perspective, we explore the potential for adapting models trained on one source domain to another. In a practical real-world scenario, assuming a model trained on one source domain (e.g., wrist data for sensor location), we adapt the model to unlabeled data from a new target domain (e.g., ankle data). Leveraging the same benchmark datasets outlined in Section \ref{sec:benchmark} and drawing inspiration from the domain adaptation framework, AdaTime \cite{ragab2023adatime}, we assess 11 unsupervised domain adaptation methods \cite{ragab2023adatime} across five types of distribution shifts and eight datasets (Table \ref{tab:da}). For each distribution shift and dataset, we conduct five experiments, randomly sampling one source and one target domain. Within each source-target experiment, five runs are performed with different random splitting seeds. Both the source and target data are divided into 75\% for training and 25\% for testing. To gauge the impact of domain adaptation, we establish an upper bound (\textit{Target Only} - model trained solely on target data) and a lower bound (\textit{No Adapt} - model trained on source data without adaptation).

We observe a performance gap on the target data between a model trained only on source data (\textit{No Adapt}) and a model trained solely on target data (\textit{Target Only}). This gap underscores the existence of a domain gap and dataset bias in the model, mirroring our observations in the domain generalization analysis. Applying unsupervised domain adaptation from one source to one target, the performance gap diminishes across all datasets and distribution shifts. Some shifts exhibit more substantial reduction than others, indicating varying degrees of shifts across different datasets. Specifically, shifts caused by sensor location, sampling rate, and device type result in a more pronounced distribution shift, with domain adaptation improving the target performance by approximately 0.11-0.2 w.r.t the lower bound \textit{No Adapt}. Conversely, user behavior shifts impact only one class in the dataset, leading to a smaller yet relatively significant drop in performance.

In comparison to domain generalization, there are noteworthy distinctions that should be noted. A key difference between domain generalization and domain adaptation, in this paper, lies not only in accessing the target domain during training but also in the utilization of multiple source domains during training. More specifically, while domain generalization draws on data from multiple related source domains to train a model that can generalize to unseen target domains, domain adaptation seeks to adapt a model trained on only one source domain to another target domain. Prior research has proposed multi-source domain adaptation methods \cite{sun2015survey, chakma2021activity}. Given the observed advantage in multi-source training with simple ERM, this analysis addresses the question of whether adapting the model from only one source is sufficient to reduce the performance gap. 

By comparing the \textit{out\_{accuracy}} performance of ERM in Table \ref{tab:dg} with the \textit{No Adapt} \textit{target\_{accuracy}} in Table \ref{tab:da}, we observe the benefit of incorporating multiple sources in the training data for certain datasets and distribution shifts without any domain generalization techniques. This benefit is evident in cases of sensor location shift in DSADS and Opportunity datasets, as well as sampling rate shifts, device type changes, and user behavior.

In contrast to the domain generalization results, where no significant improvement was observed compared to ERM for sensor location, domain adaptation enhances the \textit{target\_{accuracy}} performance, surpassing even the corresponding top performance achieved in domain generalization analysis for PAMAP2 and Opportunity. For sampling rate, the top domain generalization method for each dataset, and even simple ERM, performs slightly better than the top domain adaptation method. This suggests that incorporating data from varying sampling rates during training reduces the domain gap without the need for adaptation. Similarly, a shift in user behavior can be effectively mitigated by including multiple domains during training (ERM), with a slight improvement when applying MTL for domain generalization. However, domain adaptation is less effective in handling this domain gap, possibly due to the change being focused on one activity class. The unsupervised domain adaptation problem typically assumes that source and target domains are sampled from different marginal distributions ($P_s(x) \neq P_t(x)$), a condition not met in our juggling dataset. Conversely, shifts caused by device type exhibit additional improvement in target performance when applying domain adaptation compared to domain generalization. Similar to sampling rate and sensor location shifts, this type of shift affects the feature distribution in all activity classes. 

\subsection{Limitation and Future Work}
While this paper comprehensively explores 27 domain generalization methods and 11 domain adaptation methods across four types of distribution shifts and various datasets, it is important to acknowledge key limitations and identify potential avenues for improvement.

First, although this study is the first to delve into domain generalization and adaptation concerning specific isolated distribution shifts relevant to Human Activity Recognition (HAR), it is crucial to acknowledge that in real-world settings, data is often subject to multiple simultaneous and intricate distribution shifts. These shifts are challenging to capture and quantify in a controlled, non-real-world setting. In reality, a model needs to be capable of adapting and personalizing to a broad spectrum of changes. However, in this work, we leveraged the acknowledged challenges in model generalizability existing in the literature to simplify the problem. The objective was to understand and underscore the influence of different distribution shifts on the model's performance.     

Furthermore, despite our efforts to select a dataset that highlights a change in user behavior, it's important to note that a more longitudinal dataset would be essential to capture the full extent of distribution shifts observed in real-world scenarios. The present dataset's limitations constrain our understanding of how effectively a model can adapt or generalize to changes in behavior. However, the noticeable performance gap observed in the juggling dataset underscores the significance of studying and addressing this specific behavioral shift, particularly in the context of HAR. Future research could benefit from incorporating datasets with longer temporal perspectives to better capture real-world scenarios.

A third limitation of our study pertains to the constraints imposed on the analysis of domain generalization and domain adaptation algorithms. Our current exploration was extensive but restricted to a specific feature extractor and training implementation. For instance, the cross-entropy loss function employed in our study is not explicitly designed to handle imbalanced data. It is crucial to emphasize that our primary objective in this research was not necessarily to achieve the highest performance, but rather to highlight the challenges posed by distribution shifts and their impact on model generalizability and adaptation. Thus, model performance can be further be improved through hyperparameter tuning, experimentation with different loss functions, and exploration of other feature extractors. Our study serves as a foundational exploration of the overarching challenges, but there is room for further refinement and optimization of model performance under varying distribution shifts. Future research may consider these aspects to gain a more nuanced understanding of the capabilities and limitations of domain generalization and adaptation algorithms.

\section{Conclusion}

In this work, we shed light on the challenges faced by HAR models in generalizing to diverse sensor-relative and activity-related distribution shifts. Taking the initial steps toward a systematic evaluation, we focused on four types of sensor-based distribution shifts. Our analysis revealed that all types of shifts and datasets are primarily characterized by diversity shifts, indicating the presence of novel features not shared across domains. This highlights the complexity of the challenge faced by models in maintaining generalizability across these distribution shifts.

To assess the performance of HAR models under these conditions, we conducted a comprehensive evaluation of 27 domain generalization methods. The results indicated that existing algorithms only marginally outperform the ERM baseline on all types of distribution shifts and datasets. While some improvements were statistically significant, the overall performance advantage remains marginal, leaving considerable room for enhancement. Domain adaptation further emphasized the performance gap across domains but showcased the potential benefit of adaptation and personalization in reducing this gap.

Moreover, domain adaptation emphasizes the performance gap across domains but also demonstrates the tangible benefits of adaptation and personalization in mitigating this gap. Our benchmark platform, shared with the research community, serves as a valuable resource for testing existing methods and fostering the development of novel algorithms aimed at addressing the diverse challenges posed by distribution shifts in sensor data. This collaborative effort aims to advance the field and pave the way for more effective solutions in the realm of HAR model generalization and adaptability.

% \section{Auxiliary}

\bibliographystyle{ACM-Reference-Format}
\bibliography{bibliography}

\appendix









\end{document}